# A COMPARATIVE STUDY OF ROOT-BASED AND STEM-BASED APPROACHES FOR MEASURING THE SIMILARITY BETWEEN ARABIC WORDS FOR ARABIC TEXT MINING APPLICATIONS

Hanane FROUD[1], Abdelmonaim LACHKAR[1] and Said ALAOUI OUATIK[2]

[1] L.S.I.S, E.N.S.A,University Sidi Mohamed Ben Abdellah (USMBA),Fez, Morocco
`hanane_froud@yahoo.fr, abdelmonaime_lachkar@yahoo.fr`
[2] L.I.M, Faculty of Science Dhar EL Mahraz (FSDM), Fez, Morocco
`s_ouatik@yahoo.com`

*ABSTRACT*

*Representation of semantic information contained in the words is needed for any Arabic Text Mining applications. More precisely, the purpose is to better take into account the semantic dependencies between words expressed by the co-occurrence frequencies of these words. There have been many proposals to compute similarities between words based on their distributions in contexts. In this paper, we compare and contrast the effect of two preprocessing techniques applied to Arabic corpus: Root-based (Stemming), and Stem-based (Light Stemming) approaches for measuring the similarity between Arabic words with the well known abstractive model -Latent Semantic Analysis (LSA)- with a wide variety of distance functions and similarity measures, such as the Euclidean Distance, Cosine Similarity, Jaccard Coefficient, and the Pearson Correlation Coefficient. The obtained results show that, on the one hand, the variety of the corpus produces more accurate results; on the other hand, the Stem-based approach outperformed the Root-based one because this latter affects the words meanings.*

*KEYWORDS*

*Arabic Language; Latent Semantic Analysis (LSA); Similarity Measures; Root and Light Stemmer.*

## 1. INTRODUCTION

Arabic Documents Representation is challenge and a crucial stage; it may impact positively or negatively on the accuracy of any Text Mining tasks such as Text Categorization, Summarization, Documents Clustering, Filtering, and Retrieval purposes. Generally, Arabic Text Mining applications usually represent documents as 'Bags-of-Words' or Vector Space Model (VSM) [1][2][3], in which text documents are represented as a set of points in a high dimensional vector space. However, VSM has four primary limitations which can be grouped into two problems: the high dimensionality problem and the lack of semantics one. These limitations are: First, in information retrieval application, a long document gets a low similarity to a query because the normalized value of the document becomes high. As a result, a long-length document has little opportunity to match a query. Second, the order of words in a document is still ignored because of the bag of words assumption. The syntactic structure of a document is potentially valuable information. Third, keywords in a query have to be exactly matched with words in documents, and thus the issue of synonymy is not addressed. Fourth and finally, the issue of polysemy is not addressed because VSM only considers word form.

Therefore, for the Arabic Documents Representation not all features equally represent the document's semantics; in fact, some of these features may be redundant and add nothing to the meaning of the document; others might be synonymous and therefore capturing one of them is enough to enhance the semantic for Documents Representation purposes.

In the other hand, the Arabic Documents Representation may also be impacting by the use of different text pre-processing approaches, which affect any Text Mining tasks as we have already concluded in our previous works [7][15].

The main goal of this paper is to compare and contrast the effect of two preprocessing techniques, that affect the document's semantics, applied to Arabic corpus: Root-based (Stemming), and Stem-based (Light Stemming) approaches for measuring the semantic between Arabic words with the well known abstractive model -Latent Semantic Analysis (LSA)- with different distance functions and similarity measures [15], to overcome the above problems for Arabic Documents Representation. The LSA model is based on the Singular Value Decomposition SVD. We used SVD technique to reduce the dimensionality of the vector space [4] [5], and to build the word representative matrix. This matrix will be used later to quantify the Arabic words similarity measure. LSA technique [6] [7] is used to quantify the similarity between Arabic words by their tendency to occur in some contexts than others. The context of a word [8] consists of a set of tokens distributed on both sides of the word (after and before the word).

The rest of this paper is organized as follows. The next section introduces the concept of "Latent Semantic Analysis (LSA)" and its use for measuring similarity between two Arabic words. Section 3 describes the Stemming techniques for the Arabic Language used in the experiments. Section 4 discusses the similarity measures and their semantics. Section 5 explains experiment settings, dataset, results and analysis. In Section 6 we conclude.

## 2. LATENT SEMANTIC ANALYSIS (LSA)

Latent Semantic Analysis (LSA) is a theory and method for extracting and representing the contextual-usage meaning of words by statistical computations applied to a large corpus of text. The underlying idea is that the aggregate of all the word contexts in which a given word does and does not appear provides a set of mutual constraints that largely determines the similarity of meaning of words and sets of words to each other.

Latent Semantic Analysis (LSA) model is the automatic procedure proposed by [9] to construct a vector space. This procedure applies to a vast corpus of texts and includes the three stages below where the corpus of texts is gradually transformed into semantic vector space of several hundred dimensions. The text corpus includes two types of separators, paragraphs boundaries and spaces between words. The paragraph is regarded as the string of characters between two blanks and the word is the string of characters between two spaces.

The first step of the procedure is to represent the body as a matrix of co-occurrences. The second is to apply to this matrix a factor analysis called Singular Value Decomposition to get a space. The last step is to eliminate, among the dimensions of space resulting from the singular value decomposition, a number of dimensions, regarded as irrelevant.

### 2.1. Building the Co-occurrence Matrix

For a given corpus, the number of times each word appears in each paragraph is recognized. The frequencies of co-occurrence between words and paragraphs are calculated. These frequencies are listed in a matrix. In the Column we find each paragraph, in the row, every word. At the intersection of a column and row, each cell contains the frequency of occurrences of a word in a paragraph.

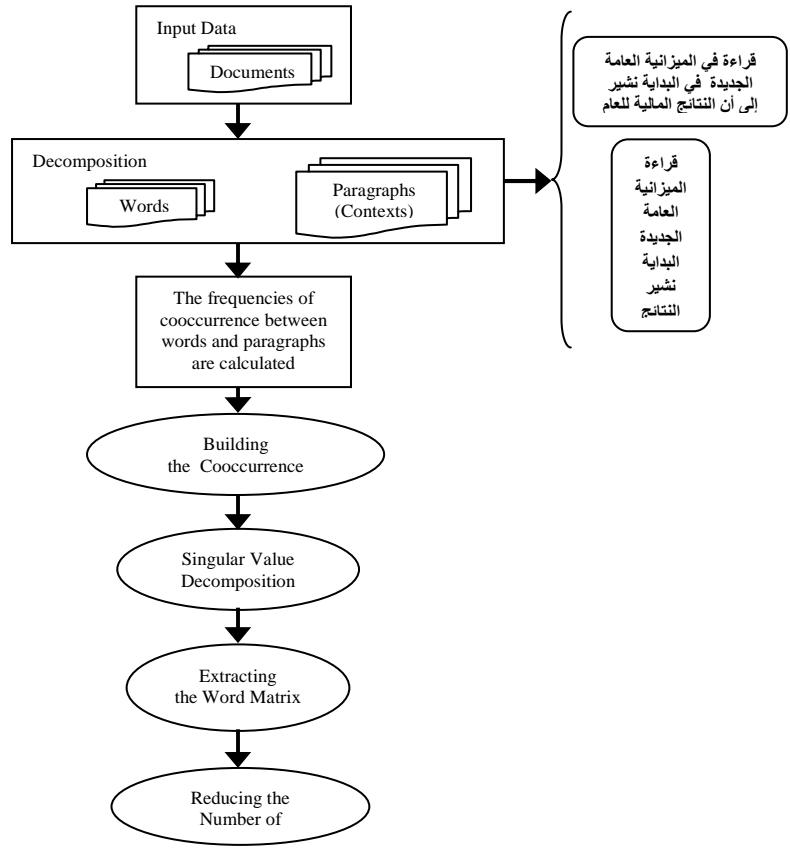

Figure 1. Latent Semantic Analysis (LSA)

## 2.2. The Singular Value Decomposition

The Singular Value Decomposition is a general method of decomposition of a linear matrix in independents principal components. As a principal components analysis, this method allows to identify a set of data - here the co-occurrence frequencies - a number of factores uncorrelated with each other and making each account for the variance of the data set. If n factors account from the totality of the variance in frequency of co-occurrence, then the data can be represented in an n-dimensional space, each dimension corresponding to a factor. The table containing the words in rows and the contexts in the columns form a rectangular matrix $X_{m*c}$, where m is the number of rows and c the number of columns. This rectangular matrix $X_{m*c}$ is decomposed into three matrices. It is the product, $U_{m*n}$, $\Delta_{n*n}$ and $V_{c*n}$:

$$X = U \Delta V^T \quad (1)$$

$\Delta_{n*n}$ matrix is a diagonal matrix with n columns and n rows, whose cells contain in the diagonal "singular values". The word matrix, U, has m lines with n values. The n values in each row are the coordinates of a vector represented in an n-dimensional space associated with a word corpus. Each word is represented in an n-dimensional space. After this step, the similarity between words can then be calculated.

## 2.3. Reducing the Number of Dimensions

All dimensions emerged from the Singular Value Decomposition is not relevant. The dimensions associated with the lowest singular values explain only a very small proportion of the variance in the original data. If these dimensions were not eliminated, the model would make errors in estimating the semantic similarity. Since the dimensions are abstract, there are no criteria for elimination of irrelevant dimensions. Consequently, the number of dimensions eliminated must be determined empirically. In most applications [6], the number of relevant factors can go from 30 000 to 300. This means that the vectors pass of 30 000 coordinates in a space of 30 000 dimensions to 300 coordinates in a space of 300 dimensions.

## 3. ARABIC TEXT PREPROCESSING

### 3.1. Arabic Language Structure

The Arabic language is the language of the Holy Quran. It is one of the six official languages of the United Nations and the mother tongue of approximately 300 million people. It is a Semitic language with 28 alphabet letters. Its writing orientation is from right-to-left. It can be classified into three types: Classical Arabic (العربية الفصحى), Modern Standard Arabic (العربية الحديثة) and Colloquial Arabic dialects (العربية العامية).

Classical Arabic is fully vowelized and it is the language of the holy Quran. Modern Standard Arabic is the official language throughout the Arab world. It is used in official documents, newspapers and magazines, in educational fields and for communication between Arabs of different nationalities. Colloquial Arabic dialects, on the other hand, are the languages spoken in the different Arab countries; the spoken forms of Arabic vary widely and each Arab country has its own dialect.

Modern Standard Arabic has a rich morphology, based on consonantal roots, which depends on vowel changes and in some cases consonantal insertions and deletions to create inflections and derivations which make morphological analysis a very complex task [22]. There is no capitalization in Arabic, which makes it hard to identify proper names, acronyms, and abbreviations.

### 3.2. Stemming

Arabic word Stemming is a technique that aim to find the lexical root or stem (Figure 2) for words in natural language, by removing affixes attached to its root, because an Arabic word can have a more complicated form with those affixes. An Arabic word can represent a phrase in English, for example the word أَتَتَذْكُرُونَنَا (ātataḏkrwnanaā[1]):"**do you remember us**?" is decomposed as follows (Table 1):

Table 1. Arabic Word Decomposition

| **Postfix** | **Suffix** | **Root** | **Prefix** | **Antefix** |
|---|---|---|---|---|
| نا | ون | تذكر | ت | أ |
| A pronoun meaning "us" | Termination of conjugation | remember | A letter meaning the tense and the person of conjugation | Preposition for asking question |

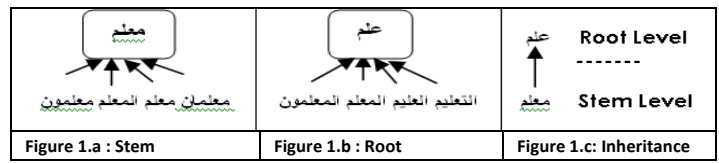

| Figure 1.a : Stem | Figure 1.b : Root | Figure 1.c: Inheritance |

Figure 2. An Example of Root/Stem Preprocessing.

---

[1] All Arabic Words in the paper are transliterated using : http://www.russki-mat.net/trans6.html

## 3.3. Root-based versus Stem-based approaches

Arabic stemming algorithms can be classified, according to the desired level of analysis, as *root-based approach* (Khoja [10]); *stem-based approach* (Larkey [11]). In this section, a brief review on the two stemming approaches for stemming Arabic Text is presented.

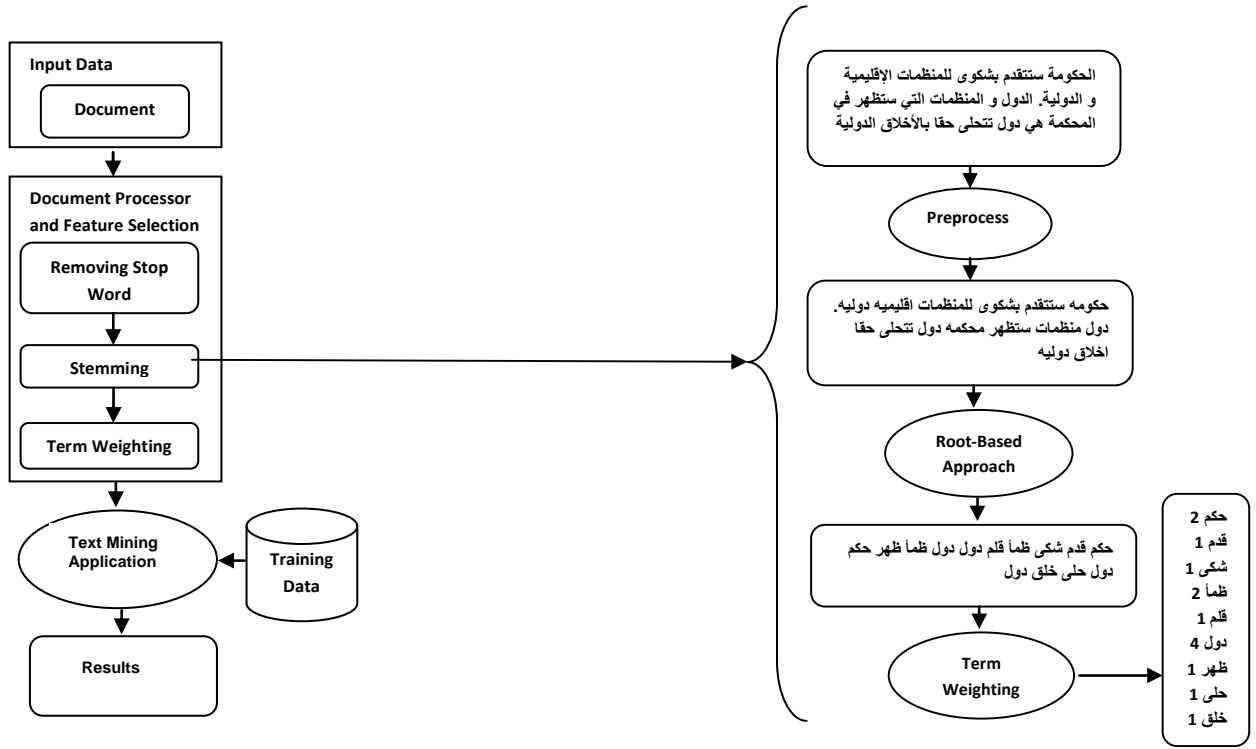

Figure 3. Example of Preprocessing with Khoja Stemmer algorithm

**Root-Based approach** uses morphological analysis to extract the root of a given Arabic word. Many algorithms have been developed for this approach. Al-Fedaghi and Al-Anzi algorithm tries to find the root of the word by matching the word with all possible patterns with all possible affixes attached to it [17]. The algorithm does not remove any prefixes or suffixes. Al-Shalabi morphology system uses different algorithms to find the roots and patterns [18]. This algorithm removes the longest possible prefix, and then extracts the root by checking the first five letters of the word. This algorithm is based on an assumption that the root must appear in the first five letters of the word. Khoja has developed an algorithm that removes prefixes and suffixes, all the time checking that it's not removing part of the root and then matches the remaining word against the patterns of the same length to extract the root [10].

The aim of the **Stem-Based** approach or Light Stemmer approach is not to produce the root of a given Arabic word, rather is to remove the most frequent suffixes and prefixes. Light stemmer is mentioned by some authors [19,20,11,21], but till now there is almost no standard algorithm for Arabic light stemming, all trials in this field were a set of rules to strip off a small set of suffixes and prefixes, also there is no definite list of these strippable affixes.

In our work, we believe that the preprocessing of Arabic Documents is challenge and crucial stage. It may impact positively or negatively on the accuracy of any Text Mining tasks; therefore the choice of the preprocessing approaches will lead by necessity to the improvement of any Text Mining tasks very greatly.

To illustrate this, in Figure 2, we show an opposite example using Khoja stemmer. It produces a root that is not related to the original word.

For example, the word (منظمات) which means (organizations) is stemmed to (ظمأ) which means (he was thirsty) instead of the correct root (نظم).

In order to test the effect of the two stemming approaches above on the similarity measures with the LSA model, we selected tow famous Stemming algorithms: **the Morphological Analyzer from Khoja and Garside [10], and the Light Stemmer developed by Larkey [11].**

## 4. SIMILARITY MEASURES

In this section we discuss the four similarity measures that were tested in [12], and we include these four measures in our work to effect the measure of the semantic between Arabic words.

### 4.1. Metric

Not every distance measure is a metric. To qualify as a metric, a measure d must satisfy the following four conditions. Let x and y be any two objects in a set and $d(x,y)$ be the distance between x and y.

1. The distance between any two points must be non-negative, that is, $d(x,y) \geq 0$.
2. The distance between two objects must be zero if and only if the two objects are identical, that is, $d(x,y) = 0$ if and only if $x = y$.
3. Distance must be symmetric, that is, distance from x to y is the same as the distance from y to x, i.e. $d(x,y) = d(y,x)$.
4. The measure must satisfy the triangle inequality, which is $d(x,z) \leq d(x,y) + d(y,z)$.

### 4.2. Euclidean Distance

Euclidean distance is widely used in clustering problems, including clustering text. It satisfies all the above four conditions and therefore is a true metric. Given two words $W_a$ and $W_b$ represented by their vectors $\vec{t}_a$ and $\vec{t}_b$ respectively, the Euclidean distance of the two words is defined as:

$$D_E(\vec{t}_a, \vec{t}_b) = \left( \sum_{t=1}^{m} \left| w_{t,a} - w_{t,b} \right|^2 \right)^{1/2}, \quad (2)$$

where $\vec{t}_a = (w_{1,a},...,w_{m,a})^T$ and $\vec{t}_b = (w_{1,b},...,w_{m,b})^T$.

### 4.3. Cosine Similarity

Cosine similarity is one of the most popular similarity measure applied to text documents, such as in numerous information retrieval applications [13] and clustering too [14]. Given two words $W_a$ and $W_b$ represented by their vectors $\vec{t}_a$ and $\vec{t}_b$ respectively their cosine similarity is:

$$SIM_C(\vec{t}_a, \vec{t}_b) = \frac{\vec{t}_a \cdot \vec{t}_b}{\left| \vec{t}_a \right| \times \left| \vec{t}_b \right|}, \quad (3)$$

where $\vec{t}_a = (w_{1,a},...,w_{m,a})^T$ and $\vec{t}_b = (w_{1,b},...,w_{m,b})^T$. The Cosine similarity varies from -1 to 1.

## 4.4. Jaccard Coefficient

The Jaccard coefficient, which is sometimes referred to as the Tanimoto coefficient, measures similarity as the intersection divided by the union of the objects. The formal definition is:

$$SIM_J(\vec{t}_a, \vec{t}_b) = \frac{\vec{t}_a \cdot \vec{t}_b}{\left|\vec{t}_a\right|^2 + \left|\vec{t}_b\right|^2 - \vec{t}_a \cdot \vec{t}_b}, \quad (4)$$

where $\vec{t}_a = (w_{1,a},...,w_{m,a})^T$ and $\vec{t}_b = (w_{1,b},...,w_{m,b})^T$.

The Jaccard coefficient is a similarity measure and it is 1 when the $\vec{t}_a = \vec{t}_b$ and 0 when $\vec{t}_a$ and $\vec{t}_b$ are disjoint.

## 4.5. Pearson Correlation Coefficient

Pearson's correlation coefficient is another measure of the extent to which two vectors are related. There are different forms of the Pearson correlation coefficient formula. Given two words $w_a$ and $w_b$ represented by their vectors $\vec{t}_a$ and $\vec{t}_b$ respectively, a commonly used form is:

$$SIM_P(\vec{t}_a, \vec{t}_b) = \frac{m\sum_{t=1}^m w_{t,a} \times w_{t,b} - TF_a \times TF_b}{\sqrt{\left[m\sum_{t=1}^m w_{t,a}^2 - TF_a^2\right]\left[m\sum_{t=1}^m w_{t,b}^2 - TF_b^2\right]}}, \quad (5)$$

where $TF_a = \sum_{t=1}^m w_{t,a}$ and $TF_b = \sum_{t=1}^m w_{t,b}$

This is also a similarity measure. However, unlike the other measures, it ranges from -1 to +1 and it is 1 when $\vec{t}_a = \vec{t}_b$.

## 5. EXPERIMENTS RESULTS AND DISCUSSION

Experiments are applied by using two Arabic testing datasets and by applying two schemes of stemming (Figure 4) : the Larkey's Stemmer developed by [11], and the Khoja's Stemmer [10]. To compute words similarity, we propose to use four schemes of different measures such as the Euclidean Distance, Cosine Similarity, Jaccard Coefficient, and the Pearson Correlation Coefficient (Figure 4).

### 5.1. Dataset

We experimented with two testing datasets both are from the website: http://www.spa.gov.sa/ for the Saudi Press Agency: the first one is a heterogeneous dataset; it's composed of 252 documents from different categories (Economics, Politics, and Sports). The second contains 257 documents belonging to one category (politics) [8]. The complete characteristics of the used corpus are described in Table.2. In the following, for each dataset we experimented with the above four similarity measures using **Larkey's Stemmer** (Table.3, Table.4) and using Khoja's **Stemmer** (Table.5, Table.6).The Euclidean Distance is a distance measure and is bounded in $[0, +\infty]$. For an ideal measure of the similarity between two Arabic words, its Euclidean distance value 0. In the other hand, the Cosine Similarity, Pearson Coefficient and Jaccard Coefficient are similarity measures and they are equal to 1 when the words are similar.

Table 2. Characteristics of the Arabic corpus used in the experiments.

| Testing Corpus | Characteristics | Value |
|---|---|---|
| Experiment n°1 (DataSet 1) | Number of Documents | 252 |
| | Size | 390 Kbytes |
| | Number of categories | 3 |
| | Number of Words | 31 321 |
| | Number of Paragraphs | 508 |
| Experiment n°2 (DataSet 2) | Number of Documents | 257 |
| | Size | 295 Kbytes |
| | Number of categories | 1 |
| | Number of Words | 23 220 |
| | Number of Paragraphs | 516 |

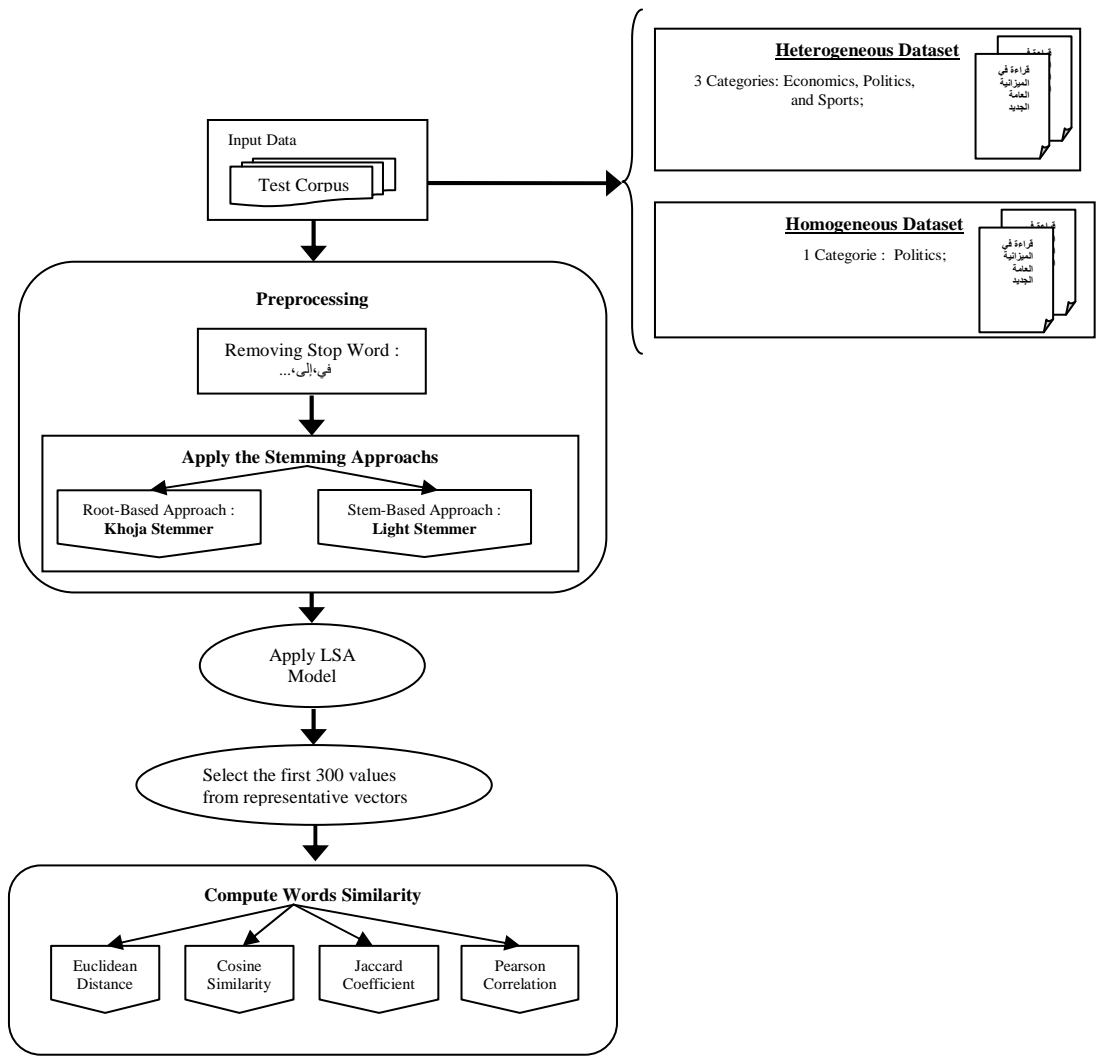

Figure 4. Description of Different Stages of Our Comparative Study

## 5.2. Results

In this work, we propose to compare the different schemes (two stemming schemes and four different measures) using both *Similar Words* and *Different Words*: ***Similar Words are*** words that have the same semantic but they may have different stems or different roots. ***Different Words are*** words that have not the same semantic even they may have the same stem or the same root.

### 5.2.1 Results with Larkey's Stemmer

*5.2.1. a. Test using Similar Words*

In Tab.3, Cosine, Pearson measures are good and the Euclidean distance is not for the pairs like: (العراقي, العراقية) and (رفضه, واستنكاره), this result shows that the similarity between those Arabic words is well detected by the LSA model using Larkey's Stemmer. Meanwhile, we can observe, in the same table, that the LSA model could not reveal the similarity between "وأوضح" and "وأفاد", "فريق" and "منتخب" (Cosine, Jaccard and Pearson measures < 0; Euclidean distance > 0).

*5.2.1. b. Test using Different Words*

In the Tab.4 we presented the obtained results using the LSA model and **Larkey's Stemmer** applied to ***Different Words***, in this table we remark that the used similarity measures performs worst for the pairs like : (البلد, الدولة), which proves that those Arabic words are not similar. But for (الرياضيين, للرياضة) the LSA model affects negatively the similarity between this words, we obtained a higher Cosine, Jaccard and Pearson measures and a worst Euclidean distances.

### 5.2.2 Results with Khoja's Stemmer

*5.2.2. a. Test using Similar Words*

The Tab.5 presents the obtained results using the four similarity measures applied to ***Similar Words*** with LSA Model using **Khoja's Stemmer**. In this case, we can observe that the results show a higher Cosine, Jaccard and Pearson measures, and a worst Euclidean distance. That prove the similarity between the pairs of words like: (العراقي, العراقية) and (أحمد, حمد). On the other hand, in the same table, the LSA Model performs worst for the pairs: (رفضه, واستنكاره) and (والدين, للشرع) that affects negatively the results obtained in Tab.3.

*5.2.2. b. Test using Different Words*

In the Tab.6, the higher Cosine, Jaccard and Pearson measures and, in the other side, the worst Euclidean distance proves that the obtained results for the pairs of Arabic words like: (السفارة, السفير) are not desirable. In the same table, the bad Cosine, Jaccard and Pearson measures show good results for the words those are not similar like: (الرئاسة, الدولة).

The above results show that the obtained results using **Larkey's Stemmer** outperformed those obtained using **Khoja's Stemmer**, because this latter affects negatively the obtained results with LSA Model [15], when we try to measure the similarity between two different Arabic words that have the same root like: (السفير, السفارة), or different root like: (رفضه, واستنكاره), (والدين, للشرع). This is mainly due to the aggressiveness of the Stemming in the sense that it reduces words to their 3-letters roots.

Table 3. Similarity Measures for Arabic Words with **Larkey's Stemmer** using LSA Model: *Using Similar Words*.

| Experiments | Words | Words Transliterated | English Translation | Cosine | Euclidean | Pearson | Jaccard |
|---|---|---|---|---|---|---|---|
| **Experiment n°1 (DataSet 1)** | (أحمد, حمد) | (ḥmd, āḥmd) | first names of people | 0.7023 | 0.1064 | 0.7034 | 0.4801 |
| | (سياسي, سياسي) | (syāsy, syāsy) | Political | 0.9999 | 6.575E-14 | 0.9999 | 0.9999 |
| | (العراقي, العراقية) | (āl'rāqy, āl'rāqy) | (Iraqi man, Iraqi women) | 1.000 | 1.327E-13 | 0.9999 | 1.000 |
| | (واستنكاره, رفضه) | (rfḍh, wāstnkārh) | Rejection | 0.6302 | 0.0433 | 0.6281 | 0.4501 |
| | (والدين, للشرع) | (llšrʿ, wāldyn) | Religion | 0.1262 | 0.1409 | 0.1266 | 0.0613 |
| | (والاستقرار, بالأمن) | (bālāmn, wālāstqrār) | (Stability, Security) | 0.0411 | 0.1442 | 0.0414 | 0.0194 |
| | (وأفاد, وأوضح) | (wāwḍḥ, wāfād) | Explain | -0.0602 | 0.1345 | -0.0601 | -0.02808 |
| | (إعانات, مساعدات) | (msā'dāt, ā'ānāt) | Aid | 0.02428 | 0.1212 | 0.0247 | 0.0111 |
| | (السنوات, الأعوام) | (ālā'wām, ālsnwāt) | Years | 0.1007 | 0.0932 | 0.0981 | 0.0339 |
| | (التطوير, التنمية) | (āltnmy, ālttwyr) | Development | 0.07793 | 0.1696 | 0.07665 | 0.04052 |
| | (فريق, منتخب) | (mntḫb, fryq) | Team | -0.0261 | 0.16017 | -0.0262 | -0.01107 |
| **Experiment n°2 (DataSet 2)** | (أحمد, حمد) | (ḥmd, āḥmd) | first names of people | 0.6073 | 0.1294 | 0.6069 | 0.41007 |
| | (سياسي, سياسي) | (syāsy, syāsy) | Political | 1.0000 | 4.631E-14 | 1.0000 | 1.0000 |
| | (العراقي, العراقية) | (āl'rāqy, āl'rāqy) | (Iraqi man, Iraqi women) | 0.9999 | 7.741E-14 | 1.0000 | 0.9999 |
| | (رفضه, واستنكاره) | (rfḍh, wāstnkārh) | Rejection | -0.0137 | 0.15069 | -0.0130 | -0.00681 |
| | (والدين, للشرع) | (llšrʿ, wāldyn) | Religion | 0.1062 | 0.20172 | 0.10232 | 0.0539 |
| | (والاستقرار, بالأمن) | (bālāmn, wālāstqrār) | (Stability, Security) | 0.01562 | 0.1521 | 0.01543 | 0.0072 |
| | (التطوير, التنمية) | (āltnmy, ālttwyr) | Development | -0.0335 | 0.1696 | -0.0343 | -0.0165 |

Table 4. Similarity Measures for Arabic Words with **Larkey's Stemmer** using LSA Model: *Using Different Words.*

| Experiments | Words | Words Transliterated | English Translation | Cosine | Euclidean | Pearson | Jaccard |
|---|---|---|---|---|---|---|---|
| **Experiment n°1 (DataSet 1)** | (السفير, السفارة) | (ālsfār, ālsfyr) | (Ambassador, Embassy) | -0.0195 | 0.17456 | -0.02273 | -0.00967 |
| | (للرياضة, الرياض) | (ālryāḍ, llryāḍ) | (Sport, Riyadh) | 0.9999 | 0.0 | 1.0 | 0.9999 |
| | (الرياضيين, للرياضة) | (llryāḍ, ālryāḍyyn) | (Sport, athletes) | 0.9999 | 0.0 | 1.0 | 0.9999 |
| | (الرئاسة, الدولة) | (āldwl, ālr'ās) | (Presidency, State) | -0.0198 | 0.1591 | -0.0202 | -0.0082 |
| | (البلد, الدولة) | (āldwl, ālbld) | (Country, State) | -0.0029 | 0.1386 | -0.0034 | -0.0013 |
| | (المجزرة, مساعد) | (msā'd, ālmǧzr) | (Assistant, Massacre) | 0.00198 | 0.1124 | 0.00112 | 7.098E-4 |
| | (اللاعبون, المنتخب) | (āllā'bwn, ālmntḫb) | (Players, Team) | 0.00860 | 0.1257 | 0.00998 | 0.004201 |
| | (السباق, احتجاج) | (āḥtǧāǧ, ālsbāq) | (Protest, Race) | 0.0165 | 0.15275 | 0.02627 | 0.00815 |
| **Experiment n°2 (DataSet 2)** | (السفير, السفارة) | (ālsfār, ālsfyr) | (Ambassador, Embassy) | -0.05302 | 0.22078 | -0.05136 | -0.02518 |
| | (الرئاسة, الدولة) | (āldwl, ālr'ās) | (Presidency, State) | -0.00415 | 0.15465 | -0.00464 | -0.00141 |
| | (البلد, الدولة) | (āldwl, ālbld) | (Country, State) | -0.00221 | 0.11529 | 1.077E-4 | -9.433E-4 |
| | (المجزرة, مساعد) | (msā'd, ālmǧzr) | (Assistant, Massacre) | 0.04222 | 0.12945 | 0.04089 | 0.01602 |
| | (للأجهزة, الاستمرار) | (ālāstmrār, llāǧhz) | (Continuation, Devices) | 0.01864 | 0.15477 | 0.018003 | 0.00941 |

This affects the semantics as several words with different meanings might have the same root (we can obtain a family of words that can be generated from the same semantic concept from a single root with different patterns).

The **Larkey's Stemmer** doesn't produce the root and therefore doesn't affect the semantics of words; but it maps several words, which have the same meaning to a common syntactical form, our observation broadly agreed with [16].

Though, for the all tables we can see that the variety of the corpus in the experiment n°1 gives more accurate results in both cases (*Similar Words* or *Different Words*).

Table 5. Similarity Measures for Arabic Words with **Khoja's Stemmer** using LSA Model:
*Using Similar Words.*

| Experiments | Words | Words Transliterated | English Translation | Cosine | Euclidean | Pearson | Jaccard |
|---|---|---|---|---|---|---|---|
| **Experiment n°1 (DataSet 1)** | (أحمد, حمد) | (ḥmd, āḥmd) | first names of people | 1.000 | 1.05E-13 | 0.999 | 1.000 |
| | (سياسي, سياسي) | (syāsy, syāsy) | Political | 1.000 | 1.39E-13 | 0.999 | 1.000 |
| | (العراقي, العراقية) | (ālʿrāqy, ālʿrāqy) | (Iraqi man, Iraqi women) | 1.000 | 0.0 | 1.0 | 1.000 |
| | (واستنكاره, رفضه) | (rfḍh, wāstnkārh) | Rejection | 0.303 | 0.063 | 0.304 | 0.177 |
| | (والدين, للشرع) | (llšrʿ, wāldyn) | Religion | 0.057 | 0.155 | 0.049 | 0.028 |
| | (والاستقرار, بالأمن) | (bālāmn, wālāstqrār) | (Stability, Security) | 0.03971 | 0.14022 | 0.0392 | 0.01949 |
| | (وأفاد, وأوضح) | (wāwḍḥ, wāfād) | Explain | -0.0237 | 0.15406 | -0.0247 | -0.016 |
| | (إعانات, مساعدات) | (msāʿdāt, āʿānāt) | Aid | 0.0061 | 0.1068 | 0.0051 | 0.00301 |
| | (السنوات, الأعوام) | (ālāʿwām, ālsnwāt) | Years | -0.0050 | 0.1070 | -0.0052 | -0.0024 |
| | (التطوير, التنمية) | (āltnmy, ālttwyr) | Development | 0.01437 | 0.1323 | 0.01735 | 0.0072 |
| | (فريق, منتخب) | (mntḫb, fryq) | Team | 0.0016 | 0.1109 | 0 0021 | 7.774E-13 |
| **Experiment n°2 (DataSet 2)** | (أحمد, حمد) | (ḥmd, āḥmd) | first names of people | 1.0 | 7.14E-14 | 1.000 | 0.999 |
| | (سياسي, سياسي) | (syāsy, syāsy) | Political | 0.999 | 3.81E-14 | 0.999 | 0.999 |
| | (العراقي, العراقية) | (ālʿrāqy, ālʿrāqy) | (Iraqi man, Iraqi women) | 1.000 | 0.0 | 1.0 | 1.000 |
| | (رفضه, واستنكاره) | (rfḍh, wāstnkārh) | Rejection | -0.086 | 0.153 | -0.089 | -0.041 |
| | (والدين, للشرع) | (llšrʿ, wāldyn) | Religion | -0.033 | 0.203 | -0.029 | -0.01 |
| | (والاستقرار, بالأمن) | (bālāmn, wālāstqrār) | (Stability, Security) | 0.00574 | 0.1239 | 0.00264 | 0.00283 |
| | (التطوير, التنمية) | (āltnmy, ālttwyr) | Development | 0.00468 | 0.17135 | 0.00473 | 0.00228 |

Table 6. Similarity Measures for Arabic Words with **Khoja's Stemmer** using LSA Model:
*Using Different Words.*

| Experiments | Words | Words Transliterated | English Translation | Cosine | Euclidean | Pearson | Jaccard |
|---|---|---|---|---|---|---|---|
| **Experiment n°1 (DataSet 1)** | (السفير, السفارة) | (ālsfār, ālsfyr) | (Ambassador, Embassy) | 1.000 | 1.52E-13 | 0.999 | 1.000 |
| | (للرياضة, الرياض) | (ālryāḍ, llryāḍ) | (Sport, Riyadh) | 1.000 | 0.0 | 1.0 | 1.000 |
| | (الرياضيين, للرياض) | (llryāḍ, ālryāḍyyn) | (Sport, athletes) | 1.000 | 0.0 | 1.0 | 1.000 |
| | (الرئاسة, الدولة) | (āldwl, ālrʾās) | (Presidency, State) | -0.001 | 0.089 | -0.002 | -5.47E-4 |
| | (البلد, الدولة) | (āldwl, ālbld) | (Country, State) | -0.005 | 0.121 | -0.005 | -0.002 |
| | (المجزرة, مساعد) | (msāʿd, ālmǧzr) | (Assistant, Massacre) | 0.004 | 0.172 | 0.006 | 0.001 |
| | (اللاعبون, المنتخب) | (ālāʿbwn, ālmntḫb) | (Players, Team) | -7.79E-4 | 0.1093 | -0.00127 | -3.831E-4 |
| | (السباق, احتجاج) | (āḥtǧāǧ, ālsbāq) | (Protest, Race) | 0.03384 | 0.11133 | 0.03238 | 0.01646 |
| **Experiment n°2 (DataSet 2)** | (السفير, السفارة) | (ālsfār, ālsfyr) | (Ambassador, Embassy) | 0.999 | 2.05E-13 | 0.999 | 0.999 |
| | (الرئاسة, الدولة) | (āldwl, ālrʾās) | (Presidency, State) | -3.68E-4 | 0.072 | -0.001 | -1.840E-4 |
| | (البلد, الدولة) | (āldwl, ālbld) | (Country, State) | -0.003 | 0.094 | -4.39E-4 | -0.001 |
| | (المجزرة, مساعد) | (msāʿd, ālmǧzr) | (Assistant, Massacre) | 0.029 | 0.224 | 0.030 | 0.013 |
| | (للأجهزة, الاستمرار) | (ālāstmrār, llāǧhz) | (Continuation, Devices) | -0.0685 | 0.2042 | -0.0683 | -0.0331 |

## 6. CONCLUSION

In this paper, we proposed to compare and contrast the effect of two preprocessing approaches: **Khoja's Stemmer** (Root-based), and **Larkey's Stemmer** (Stem-based) for measuring the similarity between Arabic words using Latent Semantic Analysis (LSA). We experimented with different schemes of similarity measures. The Root-based approach finds the three-letter roots for Arabic words without depending on any root or pattern files. The Light Stemming approach removes the common suffixes and prefixes from the words. The obtained results yield three conclusions:

1. The **Larkey's Stemmer** outperforms the **Khoja's Stemmer** because this later affects the words meanings.

2. Jaccard measure performs bad relatively to the other measures.

3. Cosine and Pearson Correlation measures, and the Euclidean Distance are quite similar for measuring the similarity between the Arabic words.

We believe that the comparative study presented in this paper will be very important; it may have a double advantage. First, it may be used generally to support the research in the field any Arabic Text Mining applications. Second, it will be used precisely to support and guides our research group to develop correctly our future works.

**Authors**

**Miss. Hanane Froud** Phd Student in Laboratory of Information Science and Systems, ECOLE NATIONALE DES SCIENCES APPLIQUÉES, University Sidi Mohamed Ben Abdellah (USMBA), Fez, Morocco. She has also presented different papers at different National and International conferences.

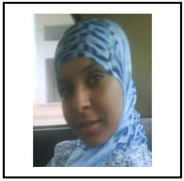

**Pr. Abdelmonaime LACHKAR** received his PhD degree from the USMBA, Morocco in 2004 in computer science; He is a Professor in the Head of Computer Science and Engineering E.N.S.A, University Sidi Mohamed Ben Abdellah (USMBA), Fez, Morocco. His current research interests include Arabic Text Mining Applications: Arabic Web Document Clustering and Categorization. Arabic Information and Retrieval Systems, Arabic Text Summarization, ect …, Image Indexing and Retrieval, 3D ShapeI Indexing and Retrieval in large 3D Objects DataBases, Colour Image Segmentation, Unsupervised clustering, Cluster Validity Index, on-line and off-line Arabic and Latin handwritten recognition, and Medical Image Applications.

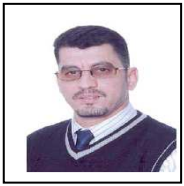

**Pr. Said Alaoui Ouatik** is working as Professor in Department of Computer Science, Faculty of Science Dhar EL Mahraz (FSDM), Fez, Morocco. His research interests include high-dimensional indexing and content-based retrieval, Arabic Document Categorization. 2D/3D Shapes Indexing and Retrieval in large 3D Objects DataBase.